\documentclass[10pt,journal,compsoc,twoside]{IEEEtran}
	\usepackage{lineno}
	\modulolinenumbers[5]

	\usepackage{amsmath}
	\usepackage{amssymb}
	\usepackage{graphicx}%
	\usepackage{dcolumn}%
	\usepackage{bm}%
	\usepackage{enumitem}
	\usepackage{stfloats}

	\newcommand\MYhyperrefoptions{bookmarks=true,bookmarksnumbered=true,	pdfpagemode={UseOutlines}, plainpages=false, pdfpagelabels=true,
	pdftitle={Seed-Point Detection of Clumped Convex Objects by Short-Range Attractive Long-Range Repulsive Particle Clustering},
	pdfsubject={Seed-point detection},
	pdfauthor={James Kapaldo},
	pdfkeywords={seed-point detection; data mining; nuclei de-clumping; short-range attractive long-range repulsive}} 

	\usepackage[colorlinks=true,allcolors=black,hypertexnames=false,\MYhyperrefoptions]{hyperref}

	\usepackage{booktabs}

	\ifCLASSOPTIONcompsoc
	  \usepackage[nocompress]{cite}
	\else
	  \usepackage{cite}
	\fi

	\renewcommand{\v}{\mathbf}
	\newcommand{\gv}{\boldsymbol}
	\newcommand{\uv}[1]{\hat{\mathbf{#1}}}

	\newcommand{\st}{\;\vert\;}
	\newcommand{\TP}{\mathrm{TP}}
	\newcommand{\FN}{\mathrm{FN}}
	\newcommand{\FP}{\mathrm{FP}}
	\newcommand{\dNDN}{\mathrm{FD_{\Delta N}}}
	\newcommand{\dN}[1]{\mathrm{FD}_{#1}}
	\newcommand{\F}[1]{F_{1,\,\delta r=#1}}

	\newcommand{\citet}[1]{\cite{#1}}

	\newcommand{\refsub}[2]{\hyperref[#1]{\ref{#1}{#2}}}

	\makeatletter
	\def\figref{\@ifnextchar[{\@with}{\@without}}
	\def\@with[#1]#2{\hyperref[#2]{{\ref{#2}#1}}}
	\def\@without#1{\hyperref[#1]{{\ref{#1}}}}
	\makeatother

	\newcommand\Tstrut{\rule{0pt}{2.4ex}}       
	\newcommand\Bstrut{\rule[-1.3ex]{0pt}{0pt}} 

\begin{document}

\title{Seed-Point Detection of Clumped Convex Objects by Short-Range Attractive Long-Range Repulsive Particle Clustering}

\author{James~Kapaldo,
		Xu~Han,
		and~Domingo~Mery,~\IEEEmembership{Member,~IEEE}
\IEEEcompsocitemizethanks{\IEEEcompsocthanksitem This work has been submitted to the IEEE for possible publication. Copyright may be transferred without notice, after which this version may no longer be accessible.
\IEEEcompsocthanksitem J. Kapaldo and X. Han are with the Department of Physics, University of Notre Dame, Notre Dame, IN, 46545. (e-mail: jkapaldo@nd.edu, xhan1@nd.edu)
\IEEEcompsocthanksitem D. Mery is with the Department of Computer Science, Pontificia Universidad Cat\'olica de Chile, Santiago 7820436, Chile. (email: dmery@ing.puc.cl)}
\thanks{Manuscript submitted Aug.\ 26, 2017.}}

\markboth{}%
{Seed-Point Detection of Clumped Convex Objects by SALR Particle Clustering}

\IEEEtitleabstractindextext{%
\begin{abstract}
	Locating the center of convex objects is important in both image processing and unsupervised machine learning/data clustering fields. The automated analysis of biological images uses both of these fields for locating cell nuclei and for discovering new biological effects or cell phenotypes. In this work, we develop a novel clustering method for locating the centers of overlapping convex objects by modeling particles that interact by a short-range attractive and long-range repulsive potential and are confined to a potential well created from the data. We apply this method to locating the centers of clumped nuclei in cultured cells, where we show that it results in a significant improvement over existing methods (8.2\% in F$_1$ score); and we apply it to unsupervised learning on a difficult data set that has rare classes without local density maxima, and show it is able to well locate cluster centers when other clustering techniques fail.
\end{abstract}

}

\maketitle
\IEEEdisplaynontitleabstractindextext


\ifCLASSOPTIONcompsoc
\IEEEraisesectionheading{\section{Introduction}\label{sec:introduction}}
\else
\section{Introduction}
\label{sec:introduction}
\fi 

	\IEEEPARstart{L}{ocating} object and distribution centers is a fundamental step of both image processing and unsupervised machine learning or exploratory data mining. The automated analysis and classification of cells from microscopy images is an important field that includes both of these disciplines \cite{Xing2016,Irshad2013,Fuchs2011,Kothari2013,Sommer2013,Navarro2016}. In this field, the first step is often locating nuclei centers using image processing techniques, which allows for seed-point based segmentation algorithms\cite{Cohen1991,Kass1988,Zimmer2002,Qi2012} to determine the nuclei boundaries. Using the segmentation results, features of each nucleus or cell are measured and used to classify the cells into different groups (e.g.\ cancer/non-cancer, different phenotypes/cell-phase\cite{Grys2017,Blasi2016,Chen2016,Sommer2013,Zhong2012,Wang2007}). When searching for new groups (new phenotypes, new effects, ...) or when it not possible for an expert to create a labeled data set, unsupervised machine learning, where groups are determined by looking for similarities in the data, must be used.

	Locating nuclei centers and using unsupervised learning to locate cluster centers are, abstractly, very similar: they both try to locate the centers of partially-overlapping convex objects or distributions. However, the techniques for solving each of these problems are quite different and are not ideal for several reasons. The current methods for locating nuclei centers (radial voting \cite{Parvin2007, Qi2012, Zhang2015, Xu2014}, gradient convergence and sliding band filters \cite{Quelhas2010, Esteves2012}, Laplacian of Gaussian filters \cite{Lindeberg1998, Al-Kofahi2010, Kong2013, Xu2016}) try to compute a surface, named the \emph{voting landscape}, that has local extrema at the nuclei centers. The fundamental problem with these methods is that the extrema of the voting landscapes are not at the nuclei centers; further, there can be extra extrema, where false nuclei are detected, or missing extrema, where true nuclei are not detected. Since the segmentation methods will return one region per seed-point, it is critical that each nuclei have only one seed-point as close as possible to the nuclei center \cite{Al-Kofahi2010,Qi2012}. For unsupervised learning, there are many well known clustering techniques (K-means/fuzzy C-means \cite{MacQueen1967,Dunn1973,bezdek1981pattern}, expectation maximization/mixture of Gaussians \cite{Dempster1977}, hierarchical clustering \cite{Johnson1967}, mean-shift \cite{Comaniciu2002}), but a general drawback of most of these methods is that they produce cluster centers that are biased towards high density regions in the data, and many of them require the number of clusters (which is normally not known beforehand) as an input. Mean-shift clustering can be better in this regard; however, each cluster must have a local maximum in the point density.

\begin{figure*}
		\centering
		\includegraphics[width=510pt]{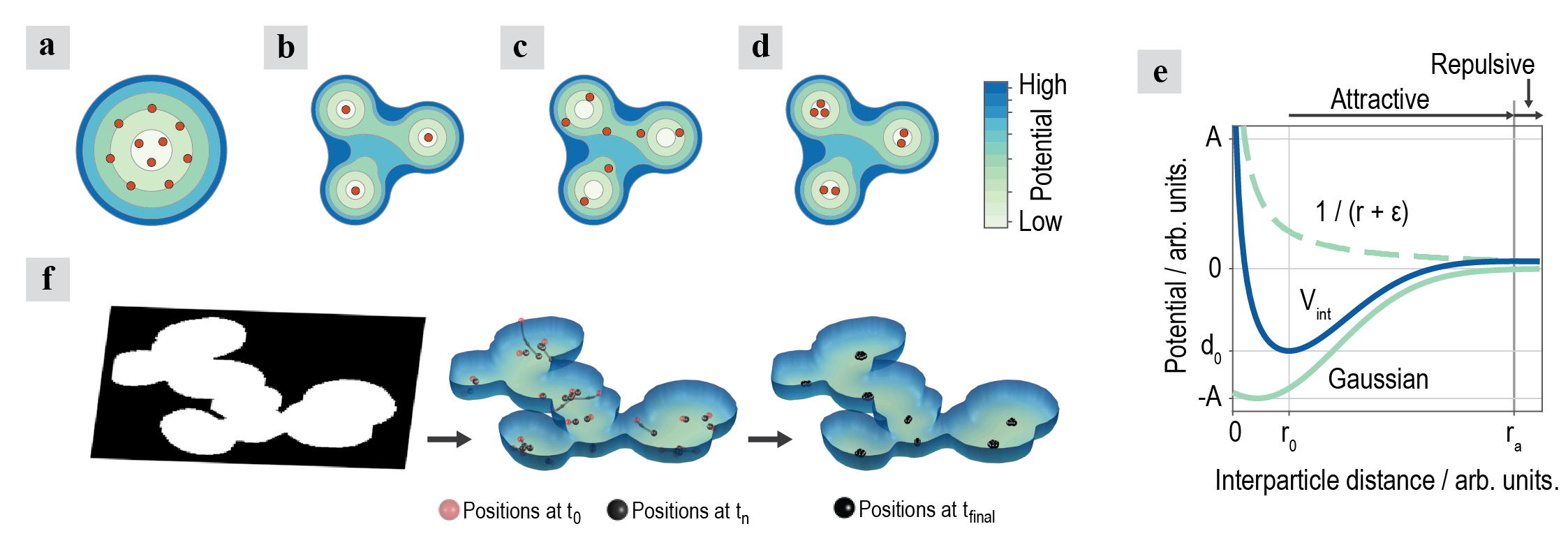}
		\caption{Premise of SALR clustering. {\bf a}--{\bf d}, Intuition: schematic of the lowest energy configuration for {\bf a}, 10 repulsive particles in a parabolic well (this is one of the two isomeric ground states\cite{Bolton1993}); {\bf b}, three repulsive particles with three potential minima; {\bf c}, seven repulsive particles with three potential minima; and {\bf d}, seven SALR interacting particles with three potential minima. Contour levels represent potential energy. {\bf e}, SALR interaction potential formed with a Gaussian attractive term and a $1/r$ repulsive term. The distance $r_a$, named the attractive extent, is the distance where particles switch from being attractive to repulsive. {\bf f}, Starting from a region of overlapping convex objects, a confining potential is created (third dimension corresponds to potential value); particles are randomly placed and the dynamics modeled; at the end the particles will form clusters located approximately at the center of each convex object. \label{fig:introduction}}
	\end{figure*}

	In this work, we develop a new clustering method (\emph{SALR clustering}) that can be used for locating both nuclei centers and the cluster centers in scatter point data, while at the same time addressing the above issues.  The premise of our method, which is graphically described in Figure~\figref{fig:introduction} and which takes cues from the fundamental physics of modeling classical Wigner crystals \cite{Bolton1993} and modeling the formation of clusters \cite{Mossa2004,Sciortino2004}, is that we can find the centers of overlapping convex regions by modeling the dynamics of particles trapped in an appropriate confining well.

	Consider a set of repulsive particles confined to a region; these particles will try to move as far apart from each other as possible. Figure~\figref[a]{fig:introduction} shows this for 10 particles confined to a parabolic potential well. If the region the particles are confined to has several local minima, then the particles will preferentially locate as close as possible to these minima to lower the total energy. Assuming our goal is to have particles near each local minima and nowhere else, we must have the same number of particles as the number of potential minima, since two particles cannot be near each other, see Figure~\figref[b,c]{fig:introduction}. This is a problem as the number of potential minima is normally not known before hand, but it can be solved by modifying how the particles interact with each other such that particles near each other are attracted to each other instead of repulsed---this short-range attractive long-range repulsive (SALR) particle interaction potential can be seen in Figure~\figref[e]{fig:introduction}. Now, we can use (many) more particles than we expect there to be potential minima, and we can have a cluster of particles located at each potential minima, see Figure~\figref[d]{fig:introduction}.

	Using this premise, we create a confining potential from the region of overlapping convex objects that has a higher energy at the region's edges and a lower energy (valley) near the object centers, we randomly place many SALR particles in this region, and then we model the particle dynamics while slowly decreasing their speed; the final position of each cluster of particles will approximately give the center of each locally convex region, see Figure~\figref[f]{fig:introduction} and {Supplemental Video 1}.

	We validate SALR clustering by applying it to the problem of locating nuclei centers, and show that it can significantly improve the location performance compared with previous methods and discuss how/why it is able to improve the performance. We also demonstrate the application of SALR clustering to unsupervised machine learning, and show that it can determine the correct number and position of clusters and better locate rare clusters that do not have local density maxima.

\section{Modeling particle dynamics}

  The equations governing the particle dynamics are derived from the Hamiltonian of a set of $N$ interacting charged particles,
  \begin{equation}
    \label{eq:hamiltonian}
    \mathcal{H} = \sum_{i}\Bigl(\frac{1}{2m_i}\v{p}_i^2 + V(\v{r}_i)\Bigr) + \sum_{\substack{i\\j\neq i}}\frac{k q_i q_j}{2}  V_\mathrm{int}\bigl(|\v{r}_i-\v{r}_j|\bigr),
  \end{equation}
  where $m$, $\v{p}$, $\v{r}$, and $q$ represent the particle's mass, momentum, position, and charge, respectfully, $k$ is a coupling constant, and $|\v{r}_i-\v{r}_j|$ is the distance between $\v{r}_i$ and $\v{r}_j$. $V_\mathrm{int}(\cdot)$ is the SALR particle interaction potential created with a Gaussian attractive term and a $1/r$ repulsive term
  \begin{equation}
    \label{eq:interactionPotential}
    V_{\mathrm{int}}(r) = \frac{1}{r + \epsilon} - A\, \exp\Biggl(-\frac{(r-\mu)^2}{2\sigma^2}\Biggr),
  \end{equation}
  where $r$ is the distance between two of the particles and $\epsilon$ is a small constant value ($\epsilon=0.2$) to prevent the divergence at $r=0$. An example of this potential is shown in Figure~\figref[e]{fig:introduction}. The values $A$, $\mu$, and $\sigma$ in \eqref{eq:interactionPotential} are set by solving a least-squares problem so that the depth $d_0$, the location of the potential minimum $r_0$, and the distance at which the potential goes from being attractive to repulsive, referred to as the attractive extent, $r_a$ may be directly specified, see Figure~\figref[e]{fig:introduction}. For example, the correspondence between two sets of these parameters $(d_0,r_0,r_a)\rightarrow(A,\mu,\sigma)$ are $(-1,2,10)\rightarrow(1.58,0.87,2.82)$ and $(-1,2,15)\rightarrow(1.84,-1.32,4.84)$.

  The particles' equations of motion can be found from the Hamiltonian \eqref{eq:hamiltonian} to be
    \begin{align}
      \frac{d\v{p}_i}{dt} &= -\gv{\nabla}_iV(\v{r}_i) - \sum_{j\neq i} k q_i q_j \gv{\nabla}_i V_\mathrm{int}(|\v{r}_i-\v{r}_j|) \\
      \label{eq:EOM_position}
      \frac{d\v{r}_i}{dt} &= \frac{1}{m_i}\v{p}_i,
    \end{align}
  where $\gv{\nabla}_i$ is the gradient operator acting on vector $\v{r}_i$.

  \begin{figure*}
    \centering
    \includegraphics[width=510pt]{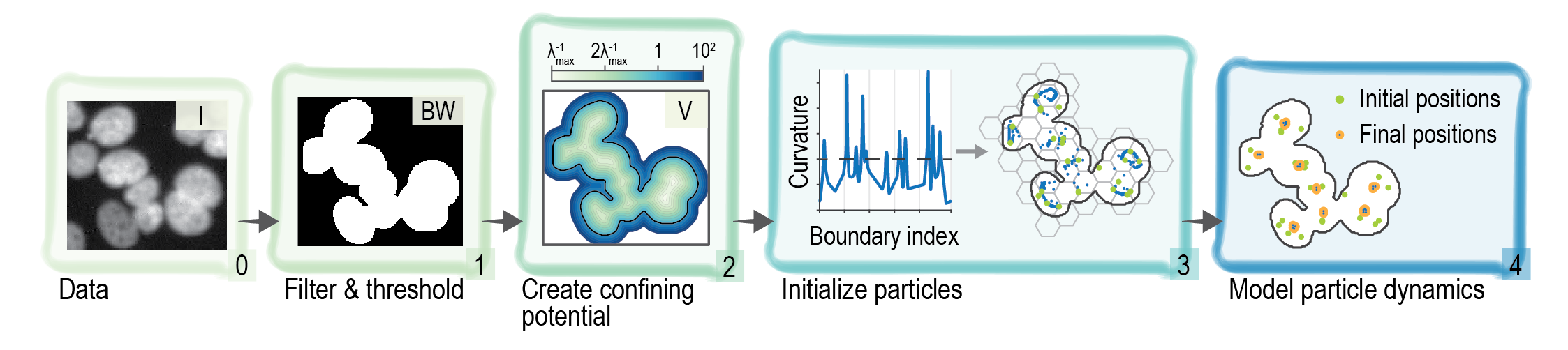}
    \caption{Locating nuclei centers method. 1) create a binary mask by filtering and thresholding the image. 2) create a confining potential (a 3D version of this potential is shown in Figure~\refsub{fig:introduction}{f}). 3) initialize particle positions: create possible set of points (blue dots) based on the boundary curvature and choose one point per grid element of an overlaid grid (green points). 4) model the particle dynamics.\label{fig:method}}
  \end{figure*} 

  The system described so far conserves energy, which means that the particles/clusters will never stop moving. To find the lowest (or a low lying) energy state of the system, we slowly decrease the temperature of the system by damping the particles:
  \begin{equation}
    \label{eq:EOM_momentum}
    \frac{d\v{p}_i}{dt} = -\gv{\nabla}_iV(\v{r}_i) -\frac{\alpha(t)}{m_i}\v{p}_i - \sum_{j\neq i} k q_i q_j \gv{\nabla}_iV_\mathrm{int}(|\v{r}_i-\v{r}_j|),
  \end{equation}
  where $\alpha(t)$ is a time dependent damping constant. $\alpha(t)$ is not required to be time dependent, but faster convergence can be achieved by increasing the value of $\alpha$ as the simulation goes on, e.g.\ $\alpha(t)\propto t$.

  Using \eqref{eq:EOM_position} and \eqref{eq:EOM_momentum} we can model the dynamics of the particles until the solution converges. We solve the set of $2N$ differential equations using a Runge-Kutta (2,3) solver\cite{Bogacki1989} (implemented with Matlab), which provides high enough tolerance to solve the problem and is faster than higher order methods. Convergence is determined by looking at the average particle speeds over a short time span (the last few time steps of the ODE solver), and if this average speed is below some threshold, then we stop the simulation. Averaging over a time span is used because the clusters of particles can have vibrational or rotational modes\cite{Blundell2011} that take a long time to die out (perhaps 30\% more computation time), and using the average speed allows these modes to be effectively ignored.

  In {Supplemental Note 1}, we make connections between our optimization method and simulated annealing and gradient descent that could lead to faster convergence.

\section{Locating nuclei centers} 

	\subsection{Method} 

	    The process we developed for applying our method to locating nuclei centers is shown in Figure~\ref{fig:method}. We go through each step below, and recommend viewing {Supplemental Video 1} for a visual description.

	    \subsubsection{Filter \& Threshold}
			After reading in the data, the (overlapping) foreground objects should be separated from the background with the goal of creating a binary mask $BW(\v{r})$ such that $BW(\v{r})=1$ for all $\v{r}$ in an object and $BW(\v{r})=0$ for all $\v{r}$ not in an object. For the nuclei images used in this work, we filtered the images with a Gaussian blur ($\sigma=1$) and then used an adaptive log-weighted Otsu threshold\cite{Otsu1979}.

	    \subsubsection{Construct confining potential}
			We use the binary mask to create a confining potential well, $V(\v{r})$. Let the set of coordinates belonging to the object be denoted by $\Omega=\{\v{r}\st BW(\v{r})=1\}$. The confining potential is given by
			\begin{equation}
			\label{eq:nucleiPotential}
			V(\v{r}) = \begin{cases} \mathrm{dt}\bigl(\neg BW\bigr)^{-1} , & \forall\; \v{r} \in \Omega \\
			          \mathrm{dt}\bigl(BW\bigr)^2+1, & \forall\; \v{r} \notin \Omega \end{cases}
			\end{equation}
			where $\mathrm{dt}(\cdot)$ is the distance transform, and $\neg$ is the negation operator ($0\leftrightarrow1$). The distance transform takes a binary image and assigns to each pixel the Euclidean distance to the nearest non-zero pixel in the binary image. The first line of \eqref{eq:nucleiPotential} states that the potential inside the object is given by 1 over the distance to the nearest object boundary pixel, and the second line states that the confining potential increases quadratically outside of the object. Note that the potential is $<1$ inside the object and $>1$ outside. After forming $V(\v{r})$, it is smoothed with a Gaussian with $\sigma=1$ pixel.

			This confining potential is not scale invariant: as a object becomes larger, the potential will become more \emph{flat-bottomed}, which will cause the particles to push each other closer to the object boundary, as is shown in {Supplemental Figure 1}a. Scale invariance can be added by implicitly scaling all objects so that their maximum distance transform values, $\lambda = \mathrm{max}(\mathrm{dt}(\neg BW))$, are equal to the same fixed value, $\lambda_\mathrm{max}$ (see {Supplemental Figure 1}b). This changes the confining potential on the interior of the object to be
			\begin{equation}
			\label{eq:scaleInvariantV}
			V_{\v{r}\,\in\,\Omega} = \biggl[1+\frac{\lambda_\mathrm{max}-1}{\lambda - 1}\Bigl(\mathrm{dt}\bigl(\neg BW\bigr)-1\Bigr)\biggr]^{-1}.
			\end{equation}
			Note that for a single ellipsoidal object, $\lambda$ represents the semi-minor axis of the object; scaling the distance transform effectively means we are resizing each object to have the same semi-minor axis, $\lambda_\mathrm{max}$.

		\subsubsection{Particle initialization}
			The next step is initializing the particles: determine the density/number of particles to use, set their mass and charge, and set their initial velocities and positions. \par

			\vspace{3pt}\textsc{Number}:
				The number of particles is determined by setting the particle density. The parameter we use to set the particle density is the Wigner-Seitz radius $r_s$, which describes the amount of space taken up by each particle. The number of particles $N$, in 2D, is then approximately
				\begin{equation}
					\label{eq:numberOfParticles}
					N=\frac{A}{\pi r_s^2},
				\end{equation}
				where $A$ is the area of the object (the number of elements in set $\Omega$) and $\pi r_s^2$ is the amount of space per particle. \par

			\vspace{3pt}\textsc{Mass \& charge}:
				We set the mass of the particles to one and the particle charge to decrease with $N$ as $q=N^{-\beta}$. Larger values of $\beta$ result in particle clusters that interact with each other less and can move around more, smaller values of $\beta$ result in particle clusters that strongly interact and push each other closer to the object boundary.\par

			\vspace{3pt}\textsc{Velocity}:
				The particles are given random initial velocities according to $\v{v}_0=s_0\uv{u}$, were $s_0$ is the initial speed and $\uv{u}$ is a random unit vector.\par

			\vspace{3pt}\textsc{Position}:
				There are several possible methods of selecting the particles' initial positions; we describe three methods which would each have their own benefit depending on the data.\par

				\vspace{3pt}\emph{i) Random}: The most simple method, choose $N$ random points out of the set $\Omega$.

				\vspace{3pt}\emph{ii) Uniform random}: To ensure we can find all of the seed-points, the particles' initial locations should uniformly cover the domain of interest ($\Omega$). This can be accomplished by overlaying a lattice across the domain and placing one particle in each lattice unit cell. In order to have approximately $N$ particles, the lattice constant is chosen such that the lattice unit cell's area is approximately the size of particle, $\pi r_s^2$ in 2D. In 2D, we use a hexagonal lattice with lattice constant $2r_s$, and in higher dimensions we use a simple (hyper)cubic lattice. The location of the particle within each lattice cell is selected by choosing one random point from the points in $\Omega$ that are also in each lattice cell, $\Omega\cap\mathrm{U}_i$, where $\mathrm{U}_i$ is the set of all coordinates in the $i$'th lattice cell.

				\vspace{3pt}\emph{iii) Convex hull transformed center of curvature (CvxHll CoC)}: Better results can be obtained if we can choose points that are approximately where the true seed-points are expected. For this, we compute a set of possible positions, $\mathrm{R}_0$, using the centers of curvature of each boundary vertex of the region $\Omega$, see {Supplemental Note 2}. For each lattice cell, we then choose one random point from the set, $\mathrm{R}_0\cap\mathrm{U}_i$. Note that the centers of curvature can be far from the geometric center of elliptical regions since the long side of an ellipse has a smaller curvature (larger radius). This is addressed by modifying the boundary curvature to be the convex hull of each negative curvature region (See {Supplemental Figure 2} for an example of this method.)

				As a final note on position initialization, the confining potential can be used to tighten the constraint on where particles can be located. At the least, we require all initial positions to have a confining potential value less than 1 (the particles must start inside the object); however, we can further require that the confining potential value at all initial positions be in some range $[V_\mathrm{min}, V_\mathrm{max}]$, which can be useful for both improving control over initial positions and reducing the number of particles used in the simulation.

	    \subsubsection{Model particle dynamics}
	      The particles are modeled using \eqref{eq:EOM_position} and \eqref{eq:EOM_momentum} until the solution converges. Box 4 in Figure~\figref[a]{fig:method} shows the particle positions after convergence; you can see clusters of particles are located at each of the nuclei centers. After the model has converged, we extract the center of each cluster by grouping together all particles within some distance of each other $(0.7\,r_0 + 0.3\,r_a)$ and then finding the mean location of each group. (This distance is empirically set and could be any distance $<r_a$ and $>r_0$.) These cluster centers are the final seed-points returned by our method.

	    \subsubsection{Reduction of work}
			It is not necessary to model the particle dynamics on all objects; if an object is convex, or it is smaller than a particle ($\pi r_s^2$), or only one particle would be modeled, then we do not run the simulation, but simply return the centroid of the object as the seed-point. We determine if an object is convex by testing that the all boundary curvature is less than some threshold, $0.25/\lambda$.

	\subsection{Results}

		\begin{figure}
		  \centering
		  \includegraphics{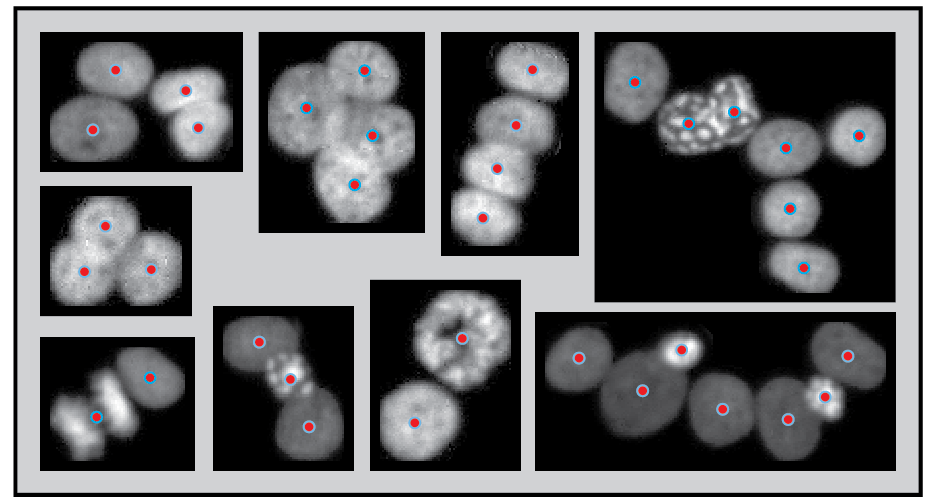}
		  \caption{Example nuclei clumps. Examples of the nuclei clumps used for validation; the contrast for each object has been enhanced. The red dots represent the labeled nuclei centers, and the edge of the markers represent our expected confidence in their location ($\sim3$ pixel radius).\label{fig:exampleImages9}}
		\end{figure} 

		\begin{figure}
		  \centering
		  \includegraphics{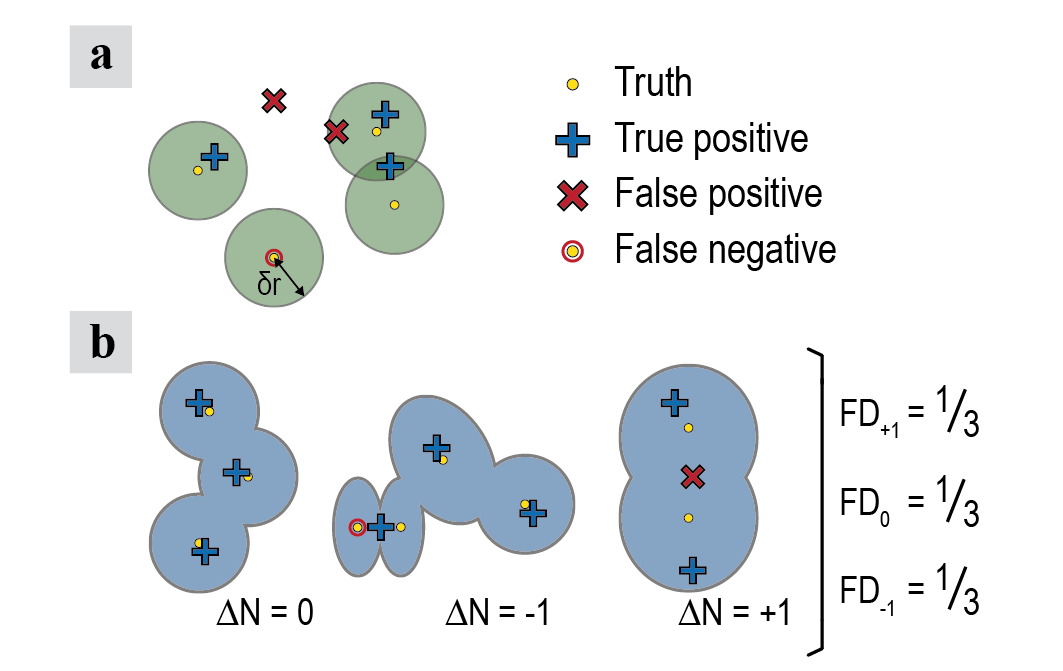}
		  \caption{Comparison metrics. {\bf a}, Definition of true positive and false positive calculated seed-points and false negative seed-points. {\bf b}, Example showing the calculation of the fractional distribution $\dNDN$ measure for three objects: one with the correct number (left), one with one too few (middle), and one with one too many (right) computed seed-points.\label{fig:comparisonMetrics}}
		\end{figure} 

		The data we use for validating our method is a set of 2,420 images of nuclei clumps with a total of 7,789 nuclei. The cells are epithelial cancer cells from the tongue (squamous cell carcinoma, SCC25) that we cultured, stained with 4',6-diamidino-2-phenylindole (DAPI) to label their DNA, and then had imaged using a whole slide fluorescence reader, see {Supplemental Note 3} for more details. The center of each of these 7,789 nuclei were manually labeled to create the truth data. Figure~\ref{fig:exampleImages9} shows an example of nine nuclei clumps with the labeled nuclei centers (more examples can be seen in {Supplemental Figure 3}).

		We use two metrics to determine the performance of the computed nuclei center locations. The first is the $F_1$ score, which ranges between 0 and 1 and is the harmonic mean of the precision and recall; a $F_1$ value of 1 means that all seed-points (nuclei centers) were calculated perfectly, and a $F_1$ value of 0 means not a single seed-point was calculated correctly. It is computed as
		\begin{equation}
			F_1 = \frac{2\,\TP}{2\,\TP + \FN + \FP}
		\end{equation}
		where $\TP$, $\FN$, and $\FP$ are the number of true positives, false negatives, and false positives, respectively. A graphical definition of these terms is shown in Figure~\figref[a]{fig:comparisonMetrics}: a calculated seed-point is $\TP$ if it is within a distance of $\delta r$ to a truth point and it is the closest calculated seed-point to that truth point, otherwise it is $\FP$; a $\FN$ point is a truth point that does not have a calculated seed-point assigned to it. For example, the rightmost red X in Figure~\figref[a]{fig:comparisonMetrics} is $\FP$ because there is another computed seed-point closer to the truth location than it.

		The second metric we use measures the fractional distribution, $\dNDN$, of objects (nuclei clumps) with the correct, too many, or too few calculated seed-points. For example, if $\dN{-1}=0.1$, then 10\% of all the objects have one less calculated seed-point than the true number. This metric is important as it directly measures the ability of a method to calculate the correct number of seed-points, even if the seed-points are not in the correct location. The fractional distribution $\dNDN$ of $\mathrm{\Delta N}$ is given by
		\begin{equation}
			\dNDN = \frac{1}{M} \sum_{i=1}^M \delta_{\mathrm{\Delta N},\:(\FP_i-\FN_i)}
		\end{equation}
		where $M$ is the number of objects (2,420), $\delta_{\cdot,\cdot}$ is the Kronecker delta, and $\FP_i$ and $\FN_i$ are the number of false positives and the number of false negatives for the $i$'th object. An example calculating $\dNDN$ is shown in Figure~\figref[b]{fig:comparisonMetrics}. Depending on how the seed-points will be used, either $F_1$ or $\dNDN$ could be more important.

		\begin{table}
			\caption{Model parameter values used for our method in computing the results of Figure \ref{fig:resultsComparison}. \label{tab:modelParameters}}
			\centering
			\begin{tabular*}{0.48\textwidth}{@{\extracolsep{\fill}}lll}
				\hline\hline
				Parameter name\Tstrut & Symbol & Default value\\
				\hline
				Particle initialization\Tstrut		& 						 & CvxHll CoC 			\\
				Wigner-Seitz radius					& $r_s$	 				 & 5 					\\
				Confining potential depth 			& $\lambda_\mathrm{max}$ & 18 		 			\\
				$V_\mathrm{int}$ attractive extent 	& $r_a$ 				 & 13 					\\
				$V_\mathrm{int}$ minimum location 	& $r_0$ 				 & 2 					\\
				$V_\mathrm{int}$ depth 				& $d_0$ 				 & -1 					\\
				Max init. particle potential 		& $V_\mathrm{max}$ 		 & 1/5 					\\
				Charge normalization 				& $\beta$				 & 1/3					\\
				Damping rate 						& $\alpha(t)$ 			 & $5\cdot10^{-4}\,t$ 	\\
				Initial speed 		 				& $s_0$					 & 0.01 				\\
				Mass								& m 					 & 1 					\\
				Coupling constant\Bstrut 			& k						 & 1 					\\
				\hline\hline
			\end{tabular*}
			\par
		\end{table}

		Using these two metrics, we compare the results of our method to seven other standard and leading methods for locating nuclei centers: multi-pass voting (MPV) by \citet{Parvin2007}, two versions of single-pass voting by \citet{Qi2012} (SPV$_\mathrm{Qi}$) and by \citet{Xu2014} (SPV$_\mathrm{Xu}$), sliding-band filter (SBF) by \citet{Quelhas2010}, generalized Laplacian of Gaussian (gLoG) by \citet{Xu2016}, maximally stable extremal regions (MSER) by \citet{Matas2002}, and directly using the local maximum of the distance transform (Dist. Trans.). We optimized the parameters of each method to maximize the sum of $\F{3}$ and $\dN{0}$; descriptions of the methods along with details on the parameters we optimized over and the final parameters used for each method can be seen in {Supplemental Note 4}. The parameters used for our method  are shown in Table~\ref{tab:modelParameters}. Many of these parameters are set empirically, and several were set by explicit optimization and are discussed in {Supplemental Note 5}.

		\begin{figure}
			\includegraphics{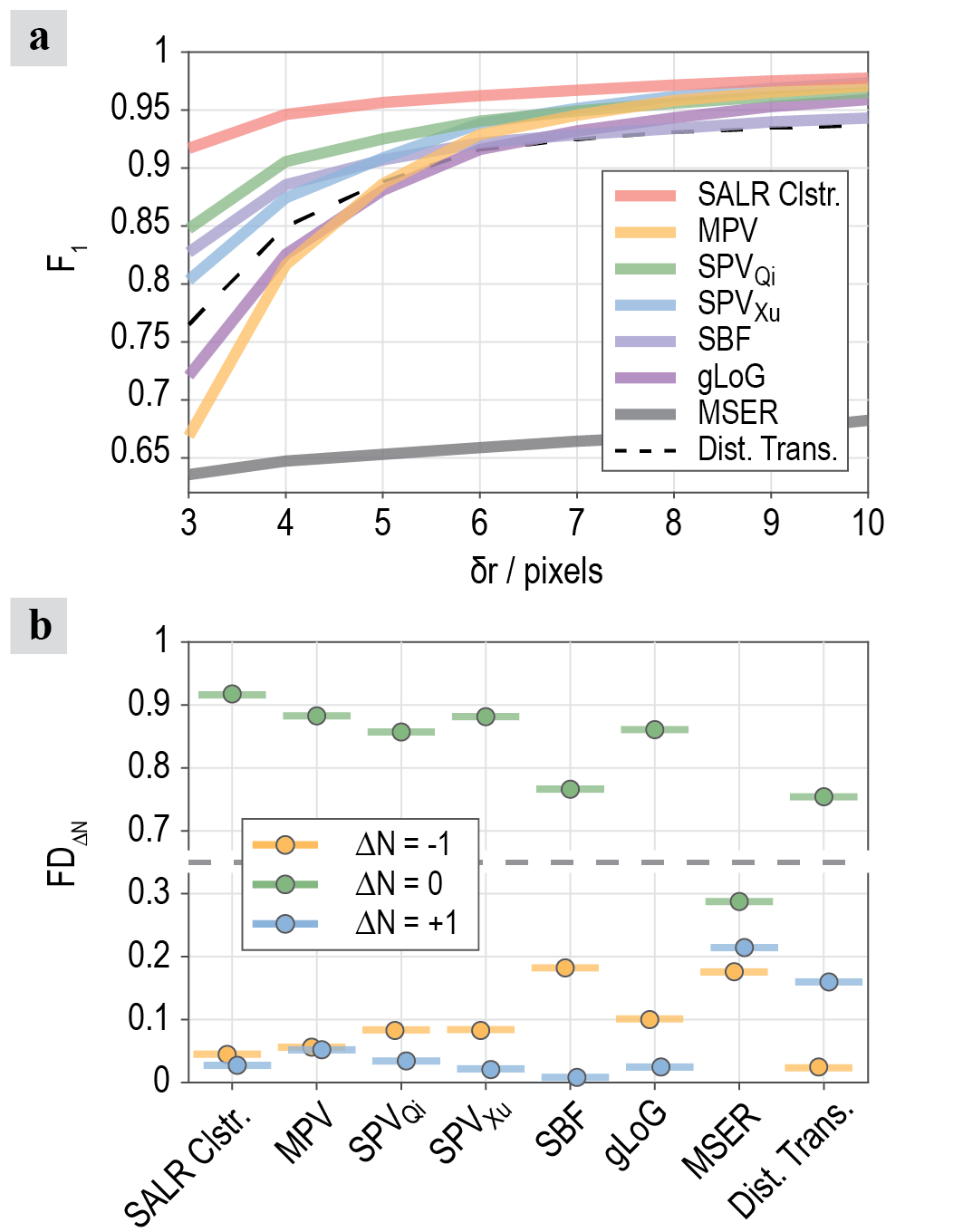}
			\caption{Methods comparison. {\bf a}, $F_1$ versus $\delta r$ for the seven methods. {\bf b}, $\dNDN$ for $\Delta N = \{-1,0,+1\}$ for the seven methods.  \label{fig:resultsComparison}}
		\end{figure} 

		We show the results of the comparison in Figure~\ref{fig:resultsComparison}. Our method (SALR Clstr.) has both the best $F_1$ score (0.069 higher, 8.2\%, than the next best method at $\F{3}$ and 0.028 higher, 3.0\%, than the next best average $F_1$ score) and the best $\dN{0}$ value (0.033 higher, 3.8\%, than the next best method).

		\begin{figure*}
			\centering
			\includegraphics[width=\textwidth]{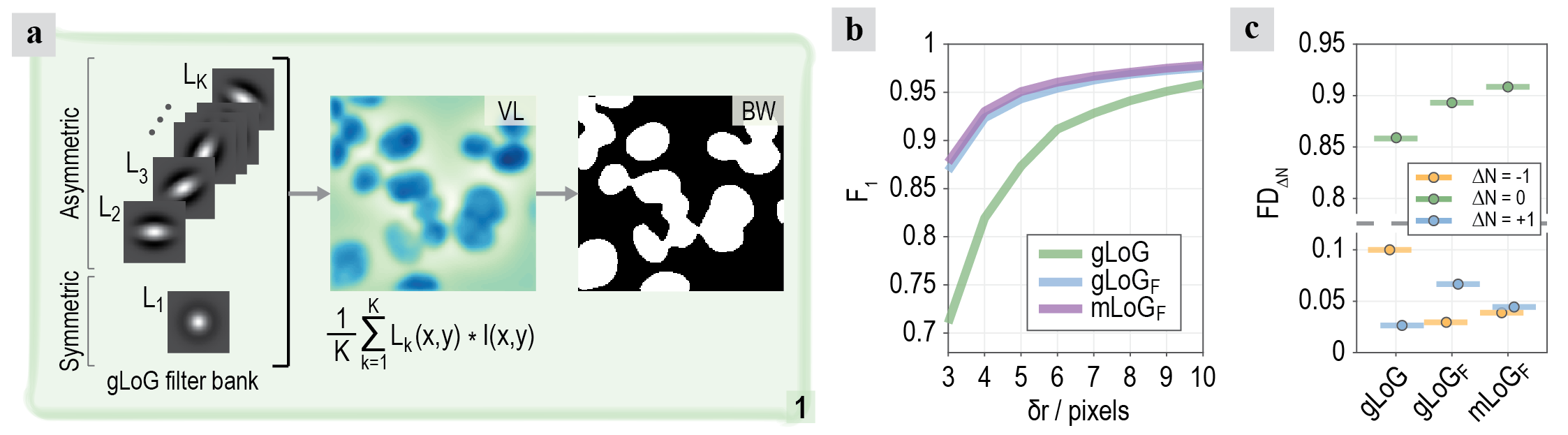}
			\caption{Advanced filtering. {\bf a}, \emph{Filter \& threshold} step of method: convolve a gLoG filter bank with an image to create a voting landscape, then threshold to create the binary mask. {\bf b}, $F_1$ score using the previous gLoG method (shown in Fig.~\ref{fig:resultsComparison}), using the entire gLoG filter bank (gLoG$_\mathrm{F}$), and using only the symmetric gLoG filter (gLoG$_\mathrm{SF}$). {\bf c}, $\dNDN$ results for the same three methods.\label{fig:gLoG_as_Filter}}
		\end{figure*} 

		In particular, note that our method does much better than the distance transform; this is important because our confining potential is proportional to one over the distance transform. This means our method is not simply locating the local maxima of the distance transform, see Discussion.

		Additional validation of our method's scale invariance as well as the performance and computation time dependence on the initial particles starting locations and density are discussed/shown in {Supplemental Note 6}/{Supplemental Figure 5} and {Supplemental Note 7}/{Supplemental Figure 6}.

	\subsection{Advanced filtering}
		The previous methods we compared to all use a voting landscape to determine the nuclei centers. We surmised that thresholding their voting landscape to create a binary mask and then using our method, instead of using the local maxima of the voting landscape, could improve the performance of the methods. We tried this using LoG filtering as it is likely the most easy to implement as well as the fastest.

		In Figure~\figref[a]{fig:gLoG_as_Filter} we show the \emph{Filter \& Threshold} step using gLoG. A gLoG filter bank is created that consists of a symmetric multiscale LoG filter (mLoG) and several asymmetric multiscale LoG filters. The voting landscape is formed by summing the convolution of each of these filters with the input image, and the binary mask is created with a simple threshold. In Figure~\figref[b]{fig:gLoG_as_Filter} and \figref[c]{fig:gLoG_as_Filter} we show the $F_1$ and $\dNDN$ results of the original gLoG method (the same results as shown in Figure~\ref{fig:resultsComparison}) and the results of using gLoG as the filter for our method (gLoG$_\mathrm{F}$). You can see that using gLoG with our method results in a significant improvement in $\F{3}$, by $\sim0.15$, and a large improvement in $\dN{0}$, by $\sim4\%$. Additionally, using only the symmetric mLoG as the filter for our method (mLoG$_\mathrm{F}$) results in even better performance than the gLoG$_\mathrm{F}$, with about a 2\% increase in $\dN{0}$.

		These results are important because much research does not use images where the nuclei can be easily segmented by thresholding (like the cultured cells in this work), but use colored histopathological (tissue) images, which are not easily thresholded. Our results show that the nuclei regions can be detected using gLoG filters and then accurate nuclei center locations and accurate nuclei count can be obtained by using our method.

\section{Application to data clustering} 
	In unsupervised machine learning and data mining, data often comes in the form of scatter point data where each point is an observation and each dimension is a different measured quantity. An example of 2D scatter point data is shown in the top of Figure~\figref[a]{fig:scatterPoint_3d}. The easiest way to apply our seed-point detection method to this type of data is by \emph{binning} the data, where a grid is overlaid on the data and the number of points in each bin is counted. The result, shown in the bottom of Figure~\figref[a]{fig:scatterPoint_3d}, can be thought of as an image, and we can threshold it to create a binary mask and then proceed as before.

	\subsection{Simple 3D data}
		\begin{figure}
			\centering
			\includegraphics{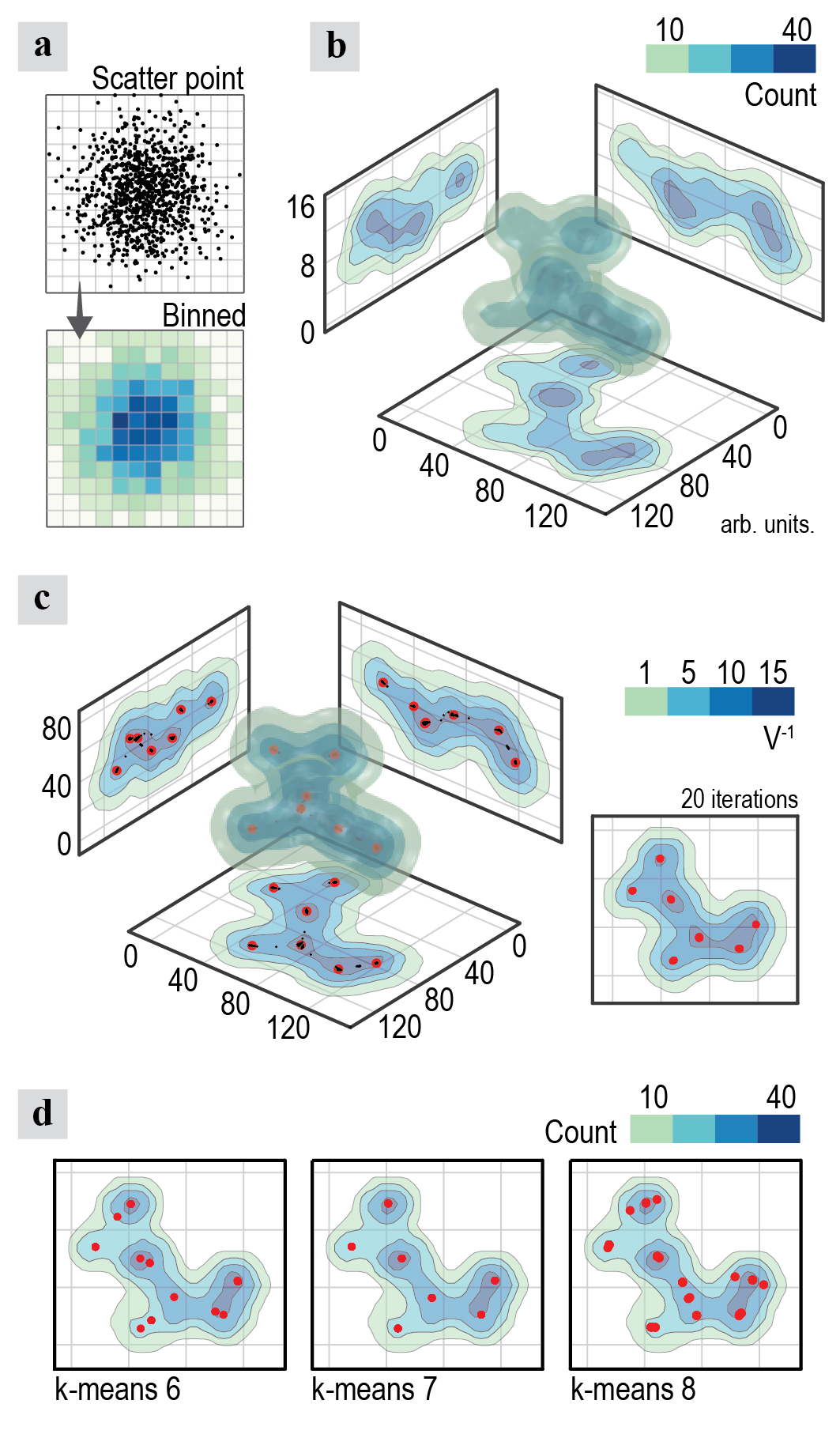}
			\caption{SALR clustering for scatter point data. {\bf a}, Example showing how scatter point data is binned. {\bf b}, Isosurface plot of the point count in our 3D example; the axis planes show the projections of the isosurfaces to 2D. {\bf c}, The confining potential created by thresholding the point count at 5 and then using \eqref{eq:scaleInvariantV}. The black points are the results of running our seed-point calculation method 20 times, and the red points are the clusters formed from these results. The 2D view of the x-y plane shows the results of repeating this process 20 times. {\bf d}, The red points show the result of running k-means clustering 20 times using 6, 7, and 8 clusters.\label{fig:scatterPoint_3d}}
		\end{figure} 

		We first validate the use of our method for detecting the cluster centers of scatter point data by using a simple 3D distribution where we can use the well known k-means clustering. The point-count isosurfaces of the 3D distribution can be seen along with the projection of the isosurfaces to the 2D axis planes in Figure~\figref[b]{fig:scatterPoint_3d}. We first scale the z-axis so that the objects are approximately the same size along any direction (see {Supplemental Note 8}), we then threshold the point-count and create the confining potential, which is shown in Figure~\figref[c]{fig:scatterPoint_3d}.

		We initialize particles for the simulation using a uniform random distribution with $r_s\approx9$ (resulting in 79 particles); all parameters other than particle density are the \emph{same} as we used in the 2D images above (values shown in Table~\ref{tab:modelParameters}). We simulated the particles 20 times (each time with new initial positions) and show the computed seed-points from each iteration as the black points in the 3D view of Figure~\figref[c]{fig:scatterPoint_3d}. The seed-points are well located in seven primary locations; the few points not at the primary locations can be reduced by tuning the particle damping rate ({Supplemental Note 9}); however, it is computationally faster to simply cluster the results of these 20 iterations and take all clusters with more than a few points as the final seed-points, which are shown as the large red dots in the 3D view. This process of clustering the results of several iterations is very robust; the 2D view in Figure~\figref[c]{fig:scatterPoint_3d} shows the results of repeating this process 20 times, and you can see the computed seed-points are all in the same location.

		We check the validity of our seven seed-points by running k-means clustering with two replicates (as implemented by Matlab's k-means++ routine) using 6, 7, and 8 clusters. We ran the clustering 20 times in each case, and show the results in Figure~\figref[d]{fig:scatterPoint_3d}. The stability of the results using 7 clusters (all 20 iterations produce the same seven positions), as compared with 6 or 8, imply that this 3D distribution does have seven clusters; and, the strong agreement in seed-point location between the k-means results and the results of our method imply that these are likely the correct locations.

		These results indicate that our method can determine the correct number of seed-points and their correct location, and, in particular, could be a good choice for locating the centers of nuclei in 3D images, which this 3D distribution resembles.

		\begin{figure*}
			\centering
			\includegraphics[width=\textwidth]{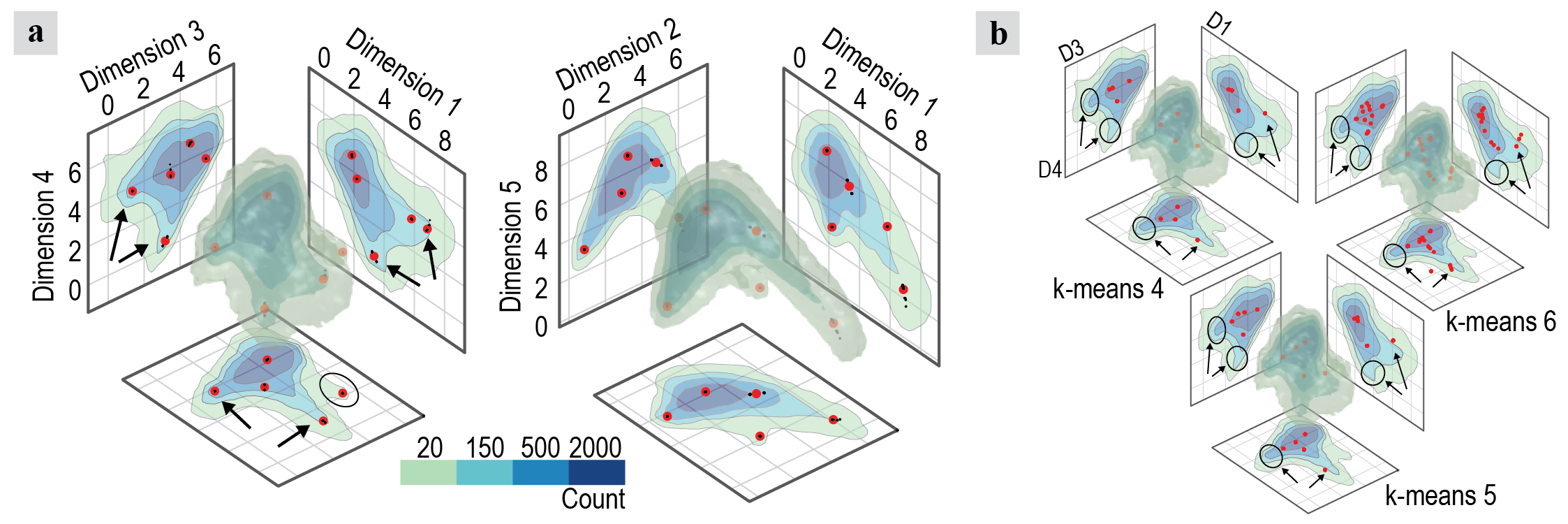}
			\caption{SALR clustering for real data. {\bf a}, projections of the 5D point count down to dimensions 1, 3, and 4 and dimensions 1, 2, and 5. Black points are the results from each of the 5 iterations, red points are the final seed-points. Arrows label the three low density regions. {\bf b}, results using k-means with 4, 5, and 6 clusters. The black arrow labels the one seed-point that approximately locates a low-density region; the circles show the two missing seed-points. \label{fig:scatterPoint_5d}}
		\end{figure*} 

	\subsection{More complex 5D data}
	    We next turn to a real data set with five dimensions and 2.7 million points; each point represents one cell' nucleus (the same SCC25 cells as above) and the five features give a measure of the damage done to the nuclei. The data is scaled in each dimension by its standard deviation and then binned (about 33 bins in each dimension). Figure~\figref[a]{fig:scatterPoint_5d} shows point-count isosurfaces for two (of the ten possible) 3D projections of the 5D data; the data has one very dense region in the center (which are undamaged nuclei) and maybe three long/thin low-density regions extending outwards from it. The long/thin regions mean that the distance transform cannot be used to create the confining potential (Supplemental Note 8). Instead, we use one over the point density for the confining potential; however, we must scale the magnitude of the potential gradient $|\gv{\nabla}V|$ so that it is approximately the same order of magnitude as the particle repulsion (see Discussion below). In addition to this change, we introduce two generalizations to our method.

		\begin{itemize}[leftmargin=*]
			\item \emph{Distance metric}: In higher dimensions, we found that we can get more stable results using the Minkowski distance metric where the distance between two points, $\v{x}$ and $\v{y}$, is defined as
			\begin{equation}
				\label{eq:Minkowski_metric}
				d = \Biggl(\sum_{i=1}^{n}|x_i-y_i|^p\Biggr)^{1/p}
			\end{equation}
			where $n$ is the dimension of the space and $p\geq1$. If we set $p=2$ then we have standard Euclidean distance; and if we set $p\to\infty$, then $d=\max_i|x_i-y_i|$, which, in our case, would mean that two particles are only attracted to each other if they are within $r_a$ in all dimensions. (Note that, in general, any distance metric can be used as long as the interaction potential is correctly defined.)

			\item \emph{Isotropic scaling}: We make a distinction between the data space (the N-D space the data is defined in) and the solver space (the N-D space we solve the equations of motion). We map between the two spaces with a simple scale factor defined to be $\ell/r_a$, where $r_a$ is the attractive extent in data space and $\ell$ is the \emph{characteristic distance} of the solver space, which, for this paper, can be directly interpreted as the attractive extent in the solver space.

			Introducing the solver space has two primary benefits: 1) If the attractive extent in data space should be $r_a\lesssim2$, then it is not possible to accurately solve for the interaction potential parameters $A$, $\mu$, and $\sigma$ (assuming $r_0\approx0.2r_a$ and $d_0=-1$). The solver space lets us scale up the problem to use an attractive extent of $\ell$ at which we can accurately solve for the parameters. 2) Changing the value of $\ell$ can precisely control the magnitude of the particle interaction (this could also be achieved using the coupling constant, $k$, if one wanted).

			In {Supplemental Figure 8}, we show how the interaction potential and force in data space change as $\ell$ goes from 10 to 50. The interaction force is strongest at $\ell=10$ and decreases as $\ell$ increases.
		\end{itemize}

		We model particles five times (small black points in Figure~\figref[a]{fig:scatterPoint_5d}) and cluster the results (red points in Figure~\figref[a]{fig:scatterPoint_5d}). You can see that our method located the three low density extrusions (labeled by black arrows) very well, and our method even finds the very small, but distinct, cluster marked by the circle in right plot of Figure~\figref[a]{fig:scatterPoint_5d}. ({Supplemental Video 2} shows an animation of the particle dynamics along with the comparison to k-means with 5 clusters.) The parameter values used in these simulations were as follows; uniform random distribution, $r_s\approx1.5$, $[V_\mathrm{min},V_\mathrm{max}]=[1/4,1/6]$, $r_a=2$, $r_0=0.15\,r_a$, $d_0=-1$, $\ell=12$, $p=4$, and the potential force was scaled so that the 99\% value is 0.4. (The potential bound requirements helps to reduce the number of particles, which is 147, and to position the particles near the outskirts of the region, which is better, in general.)

		As means of comparison, we used k-means clustering with two repetitions for 4, 5, and 6 clusters; we repeated this 20 times and show the results in Figure~\figref[b]{fig:scatterPoint_5d}. The results using 4 or 5 clusters are stable (the clusters are in the same place in all 20 iterations), but are only able to approximately locate one of the three low density regions, while missing two of them (labeled by the circles). The reason k-means does not locate the low density regions well is that the cluster centers it finds are the locations which minimize the variance in each cluster, and the variance can be minimized by moving the cluster centers closer to the high density regions. (The same/similar reasoning also applies for fuzzy c-means or mixture of Gaussians.) This demonstrates an important feature of our method: our method can locate the center of low density regions (also called rare classes) even when the low density regions do not have local maximum.

		Looking at the computation time, the k-means with 5 clusters, which was the fastest, takes $15\pm5$ sec per repetition; our method takes $1.15\pm0.08$ sec per iteration with $\sim5.5$ sec overhead for binning, smoothing, and computing the confining potential gradient. (Note each repetition of k-means and each iteration of our method can run in parallel. Computer specifications are shown below.) Thus, our method can lead to an important speed increase for some data sets.

\section{Discussion} \label{sec:discussion}

	It is interesting to consider how, and under what conditions, SALR clustering leads to these improvements. The answer comes by analyzing the force on the particles in the simulation: the confining potential exerts a force on the particles with a magnitude of $|\gv{\nabla}V|$ (which is the slope of the confining potential in Figure~\figref[f]{fig:introduction}), and the particles repel each other with a force of about $r_a^{-2}$. If the force from the confining potential is much larger than the repulsive force, $|\gv{\nabla}V|\ggg r_a^{-2}$, then the particles will only find the local minimum of the confining potential---this would lead to the results labeled Dist.\ Trans.\ in Figure~\ref{fig:resultsComparison}, and is the same thing that the previous methods do in finding the local extrema of the voting landscape or fining the local maxima of the point density (mean-shift). Alternatively, if the repulsive force is much larger than the confining force, $r_a^{-2}\ggg|\gv{\nabla}V|$, then the particles will ignore the confining potential minima and spread out---this would result in particle clusters separated by at least a distance $r_a$ that uniformly cover the region. In between these two regimes, when the two forces are approximately the same order of magnitude, $|\gv{\nabla}V|\sim r_a^{-2}$, there is an interaction between the confining potential and the particle repulsion, and this is the regime we operate in and that leads to SALR clustering's improved performance. (When using the distance transform to create the confining potential in our images, our confining force is $\propto r^{-2}$, where $r$ is the distance to the nearest region boundary, and the repulsive force between the particles is also $\propto r^{-2}$, where $r$ is the distance between particles.) This need for the forces to be approximately equal is why it is necessary to scale the magnitude of the potential gradient when the inverse point density is used as the confining potential.

	Our method can be applied to even higher dimensional data sets, but it likely will not be possible to bin the data as we did above. Binning the data requires a large amount of space (the 5D data set, 37x30x28x34x36, takes $\sim150$ MB as single precision), and is not necessary when the density is used as the confining potential. The benefit of binning the data is computational speed, as it allows us to pre-compute the density and its gradient everywhere. If the data cannot be binned, then the gradient of the density will need to be calculated (using a nearest neighbors range search) while solving the differential equations; this could be quite slow. Additionally, the magnitude of the gradient should be scaled; thus, before modeling the particle dynamics, a random sampling of the gradient magnitude will need to be performed so that the scale factor can be determined.

	There are still many aspects of SALR clustering that are missing or can be improved and expanded upon, such as using an asymmetric particle interaction, developing a performance metric, implementing a clustering method (where each scatter point is placed into a specific cluster), and using multiple confining potentials and types of particles. These are each briefly discussed in {Supplemental Note 10}.

\section{Conclusion}
	SALR clustering can represent a significant improvement in locating the centers of overlapping convex objects: it locates the correct number of nuclei more often and the nuclei centers more accurately than standard and leading methods; it can significantly improve the performance of previous methods; and it is able to determine, not only the number of clusters, but the correct position of the cluster centers in data clustering while not requiring a cluster to have a local density maximum.

\ifCLASSOPTIONcompsoc
  \section*{Acknowledgments}
\else
  \section*{Acknowledgment}
\fi
	The authors would like to acknowledge Guillermina Garcia from Sanford-Burnham Medical Research Institute for providing the whole slide fluorescence imaging. J.K.\ acknowledges Prof. Sylwia Ptasinska for the opportunity and resources to work on this project. J.K.\ and X.H.\ acknowledge the support from the US Department of Energy Office of Science, Office of Basic Energy Sciences, under Award Number DE-FC02-04ER15533. This is contribution number NDRL 5175 from the Notre Dame Radiation Laboratory.

\section*{Computer specifications}
	Computation times related to locating the nuclei centers are based on using a 64-bit i7-4790 CPU \@ 3.60 GHz with 32.0 GB ram. Computations times reported for the 5D data set are based on using a 64-bit i7-3770 CPU \@ 3.40 GHz with 12.0 GB ram.

\section*{Code \& data availability}
	The SALR clustering code (written in Matlab) and documentation, as well as all images, truth data, and scatter point data used in this work, are in {Supplementary Software and Data} and available on GitHub (\url{https://github.com/jkpld/SALR_Clustering}). In particular, we include example scripts that directly reproduce the results of this work.

\section*{Author contributions}
	J.K.\ conceived the project, implemented the method, created validation truth data, analyzed the results, and wrote the manuscript. X.H.\ performed all tasks relating to the cell work (cell culture, cell experiments, staining the cells, having them imaged) and provided helpful discussion throughout the course of the project. D.M.\ provided valuable support and feedback in elevating the project and improving the manuscript.

	\bibliographystyle{IEEEtran}
	\bibliography{bibliography}

\newpage

\begin{IEEEbiography}[{\includegraphics[width=1in,height=1.25in,clip,keepaspectratio]{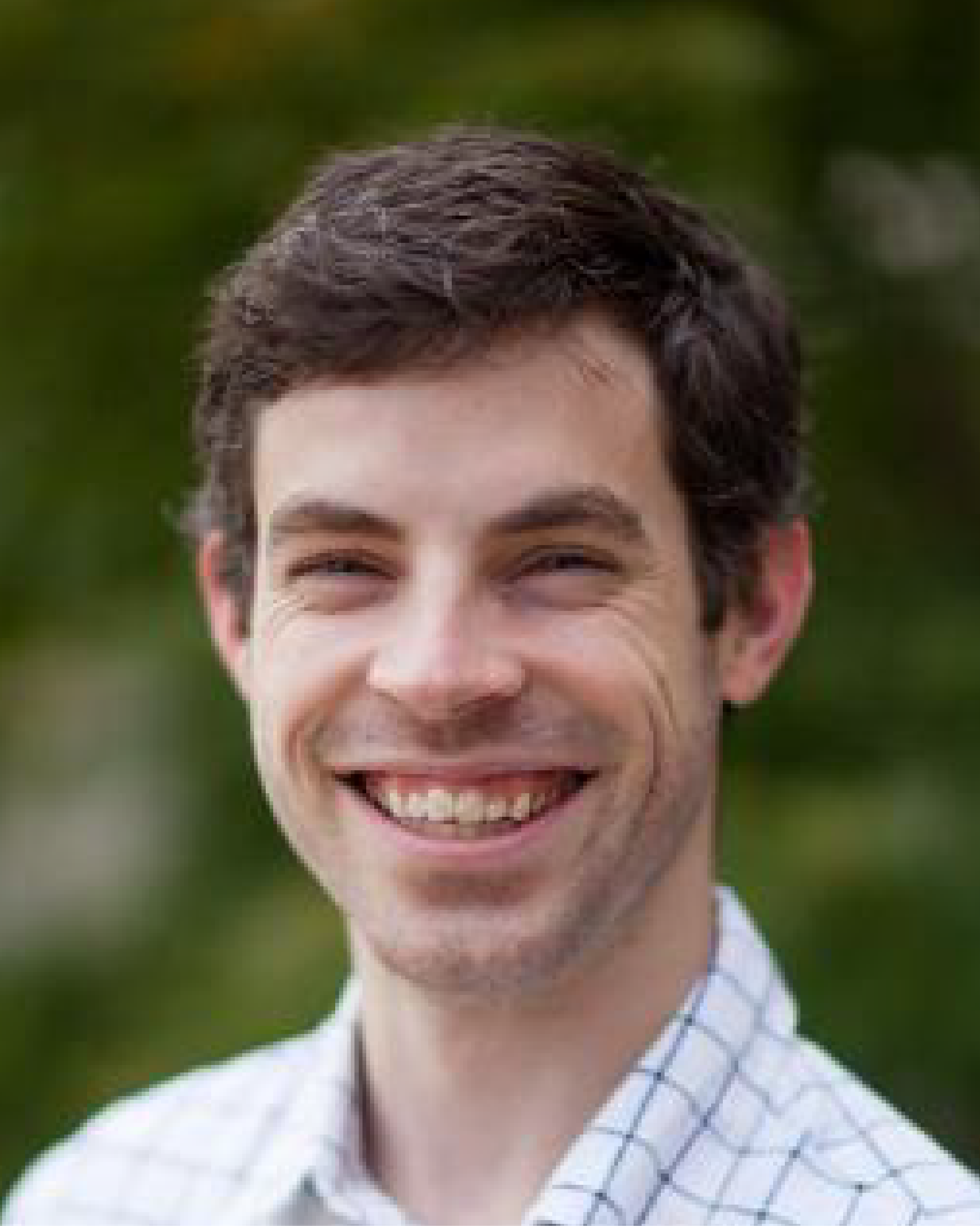}}]{James Kapaldo}
James Kapaldo is a Ph.D candidate in Physics at the University of Notre Dame. His current research interests include machine learning and its use for analyzing scientific experiments, experimental design and automation, the study of cold plasma jets and their effect on oral cancer cells. Kapaldo received a M.Sc. degree in physics from the University of Notre Dame studying Wigner Molecules and Ga-In intermixing in InP/GaInP quantum dots. During the summers '09 (on a DAAD fellowship) and '10 he was a research intern at Max Planck for Quantum Optics near Garching, Germany where he worked on designing and building a ultra-high vacuum beamline for attosecond laser experiments. Contact him at jkapaldo@nd.edu.
\end{IEEEbiography}

\begin{IEEEbiography}[{\includegraphics[width=1in,height=1.25in,clip,keepaspectratio]{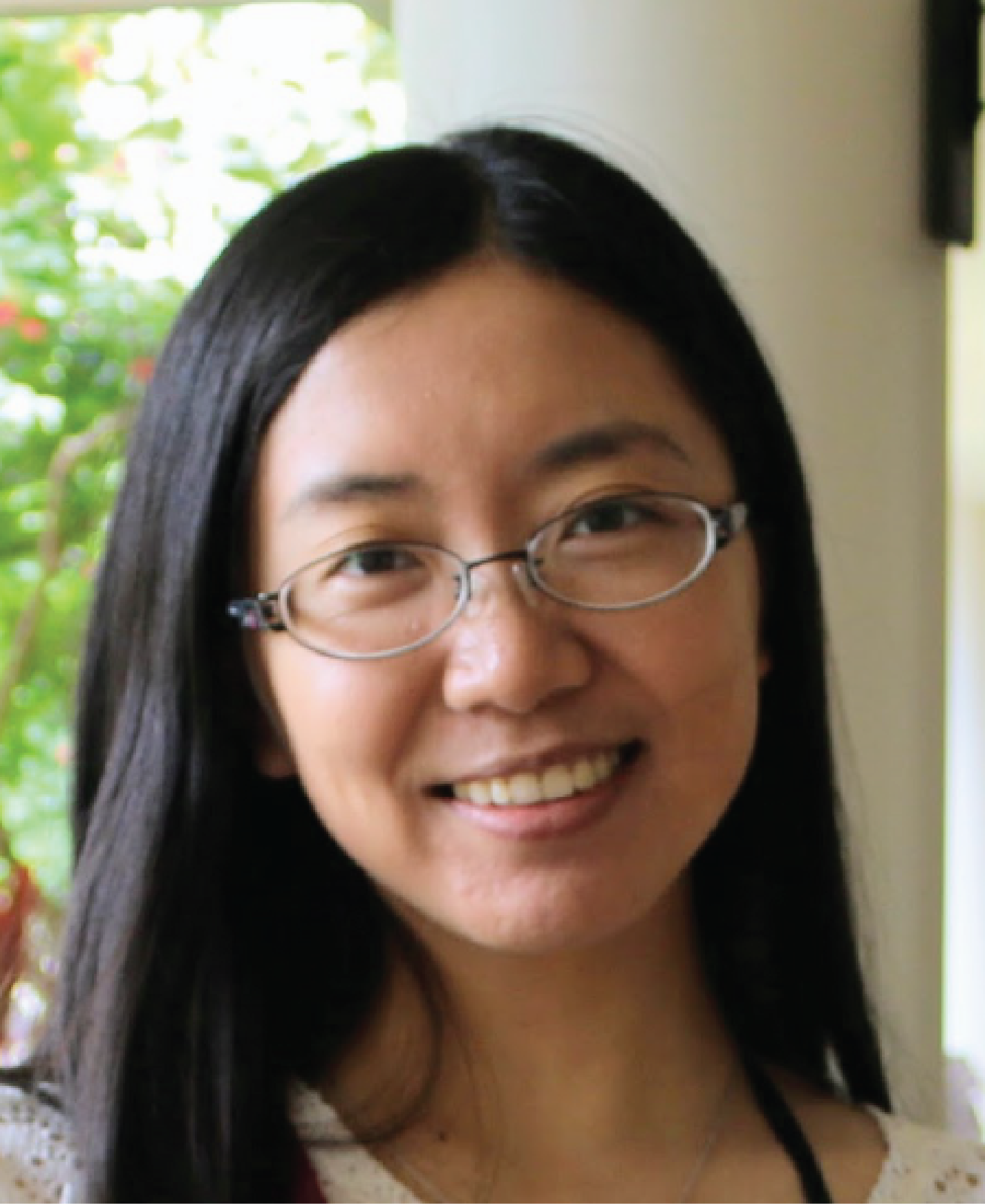}}]{Xu Han}
Xu Han received a Ph.D ('17) in Physics at the University of Notre Dame. Her research interests include studying the effect of cold plasma jets on oral cancer cells, using dosimetry methods for studying the chemical changes induced by cold plasma jets, and cold plasma jet diagnostics. Contact her at xhan1@nd.edu.
\end{IEEEbiography}

\begin{IEEEbiography}[{\includegraphics[width=1in,height=1.25in,clip,keepaspectratio]{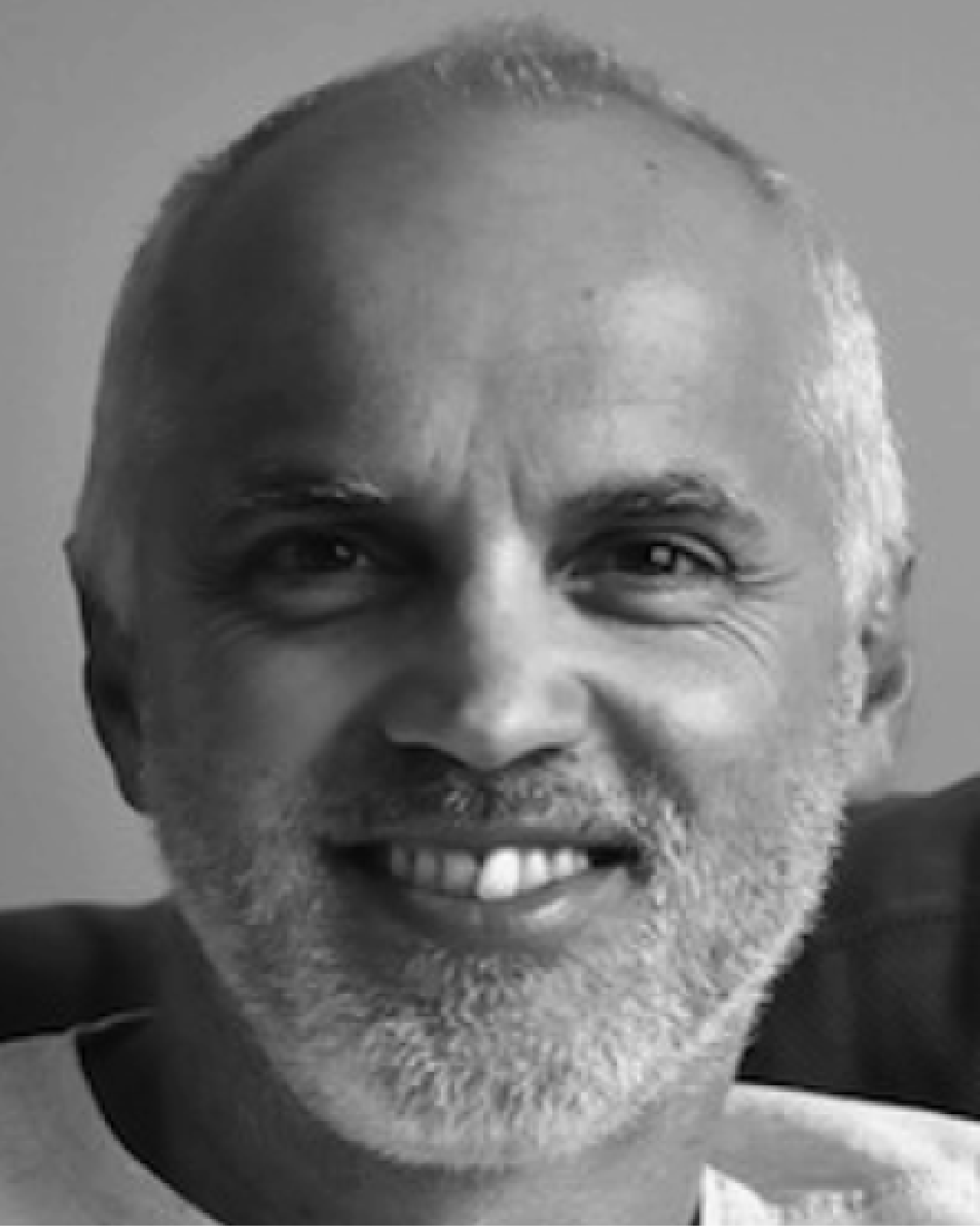}}]{Domingo Mery}
Domingo Mery (M’01) received the M.Sc. degree in electrical engineering from the Technical University of Karlsruhe in 1992, and the Ph.D. degree (Hons.) from the Technical University of Berlin in 2000. He was a Research Scientist with the Institute for Measurement and Automation Technology, Technical University of Berlin, in collaboration with YXLON X-Ray International. In 2001, he served as an Associate Researcher with the Department of Computer Engineering, Universidad de Santiago, Chile. In 2014, he was a Visiting Professor with the University of Notre Dame. He is currently a Full Professor with the Department of Computer Science, Pontificia Universidad Católica de Chile, where he served as the Chair from 2005 to 2009. He has authored or coauthored 60 technical SCI publications and over 70 conference papers. His research interests include image processing for fault detection in aluminum castings, X-ray imaging, real-time programming, and computer vision. He has received scholarships from the Konrad Adenauer Foundation and the German Academic Exchange Service. He received the Ron Halmshaw Award in 2005 and 2012, the John Green Award from the British Institute of Non-destructive Testing in 2013, which was established to recognize the best papers published in the Insight Journal on Industrial Radiography, and the Best Paper Award at the International Workshop on Biometrics in conjunction with the European Conference on Computer Vision in 2014. He is a Local Co-Chair of ICCV2015 (to be held in Santiago de Chile). He served as the General Program Chair of PSIVT2007, the Program Chair of PSIVT2009, and the General Co-Chair of PSIVT2011 (Pacific-Rim Symposium on Image and Video Technology) and the 2007 Ibero-American Congress on Pattern Recognition.
\end{IEEEbiography}

\vfill


%



\end{document}